\documentclass[letterpaper]{article} 
\usepackage{aaai25}  
\usepackage{times}  
\usepackage{helvet}  
\usepackage{courier}  
\usepackage[hyphens]{url}  
\usepackage{graphicx} 
\usepackage{natbib}  
\usepackage{caption} 
\usepackage{algorithm}
\usepackage{algorithmic}

\usepackage{newfloat}
\usepackage{listings}
\DeclareCaptionStyle{ruled}{labelfont=normalfont,labelsep=colon,strut=off} 
\floatstyle{ruled}
\newfloat{listing}{tb}{lst}{}
\floatname{listing}{Listing}
\usepackage{booktabs}       
\usepackage{amsfonts}       
\usepackage{xcolor}         
\usepackage{amssymb}
\usepackage{makecell}

\usepackage{pifont}
\newcommand{\cmark}{\ding{51}}
\newcommand{\xmark}{\ding{55}}
\usepackage{arydshln}
\usepackage{multirow}
\usepackage{amsmath}

\title{Personalized Lip Reading: \\ Adapting to Your Unique Lip Movements with Vision and Language}
\author{
    \\
    Jeong Hun Yeo\textsuperscript{\rm 1},
    Chae Won Kim\textsuperscript{\rm 1},
    Hyunjun Kim\textsuperscript{\rm 1},\\
    Hyeongseop Rha\textsuperscript{\rm 1},
    Seunghee Han\textsuperscript{\rm 1},
    Wen-Huang Cheng\textsuperscript{\rm 2},
    Yong Man Ro \textsuperscript{\rm 1}\thanks{Corresponding author}
}
\affiliations{
    \textsuperscript{\rm 1}Integrated Vision and Language Lab, KAIST, South Korea\\
    \textsuperscript{\rm 2}National Taiwan University, Taipei, Taiwan  \\
    sedne246@kaist.ac.kr, chaewonkim@kaist.ac.kr, kimhj709@kaist.ac.kr, \\ ryool\_1832@kaist.ac.kr, gkstmdgml211@kaist.ac.kr, wenhuang@csie.ntu.edu.tw, ymro@kaist.ac.kr

}

\begin{document}

\maketitle

\begin{abstract}
Lip reading aims to predict spoken language by analyzing lip movements. Despite advancements in lip reading technologies, performance degrades when models are applied to unseen speakers due to their sensitivity to variations in visual information such as lip appearances. To address this challenge, speaker adaptive lip reading technologies have advanced by focusing on effectively adapting a lip reading model to target speakers in the visual modality. However, the effectiveness of adapting language information, such as vocabulary choice, of the target speaker has not been explored in previous works. Additionally, existing datasets for speaker adaptation have limited vocabulary sizes and pose variations, which restrict the validation of previous speaker-adaptive methods in real-world scenarios. To address these issues, we propose a novel speaker-adaptive lip reading method that adapts a pre-trained model to target speakers at both vision and language levels. Specifically, we integrate prompt tuning and the LoRA approach, applying them to a pre-trained lip reading model to effectively adapt the model to target speakers. Furthermore, to validate its effectiveness in real-world scenarios, we introduce a new dataset, VoxLRS-SA, derived from VoxCeleb2 and LRS3. It contains a vocabulary of approximately 100K words, offers diverse pose variations, and enables the validation of adaptation methods in the wild, sentence-level lip reading for the first time in English. Through various experiments, we demonstrate that the existing speaker-adaptive method also improves performance in the wild at the sentence level. Moreover, we show that the proposed method achieves larger improvements compared to the previous works.
\end{abstract}

%
\begin{links}
    \link{Code}{https://bit.ly/Personalized-Lip-Reading}
    \link{Datasets}{https://bit.ly/VoxLRS-SA-Dataset}
    \link{Extended version}{https://bit.ly/2409-00986}
\end{links}

\section{Introduction}
Lip reading, also known as Visual Speech Recognition (VSR), aims to predict spoken language by analyzing visual cues from lip movements. It plays a crucial role in enhancing human communication in bustling cafes, crowded streets, or loud factories where audio signals are compromised. Furthermore, it can support individuals with hearing impairments, enabling them to comprehend speech through visual means. With these practical benefits and the development of deep learning, we can accurately infer what a speaker says without a speaker's voice by utilizing recent lip reading systems \cite{shi2022learning, ma2023auto, yeo2024akvsr}.

Despite significant advancements in lip reading technologies, their performance often declines when applied to unseen speakers not included in training data \cite{assael2016lipnet, kim2022speaker}. This limitation arises because lip reading models are inherently sensitive to variations in lip appearances, movements, and speaking speeds among different individuals. To address this issue, recent studies \cite{kim2023prompt, he2024speaker, luo2023learning, wu2024landmark} have incorporated speaker adaptation techniques into lip reading. These techniques aim to adapt pre-trained lip reading models to unseen speakers by utilizing minimal additional training parameters. Specifically, these works have applied prompt tuning \cite{brown2020language, zhou2022conditional} and Low-Rank Adaptation (LoRA) \cite{hu2021lora} strategies to improve the adaptability and performance of lip reading models on target speakers.


\begin{table*}[t]
  \renewcommand{\arraystretch}{1.1}
  \renewcommand{\tabcolsep}{2.5mm}
  \centering
  \resizebox{0.95\linewidth}{!}{
  \begin{tabular}{cccccccc}
    \Xhline{3\arrayrulewidth}
    \textbf{Dataset} & \textbf{Segment} & \textbf{Split} & \textbf{\# speakers} &  \makecell{\textbf{Duration (hrs)}}  & \textbf{\# words} & \makecell{\textbf{Pose \& Background} \\ \textbf{Diversity}} & \makecell{\textbf{Transcription} \\ \textbf{labels}}
    \\ \hline
    \multirow{2}{*}{\makecell{\textbf{GRID} \\ \cite{cooke2006audio}}}
    & \multirow{2}{*}{\makecell{\textbf{Baseline} \\ \textbf{\& Adaptation}}} & \multirow{2}{*}{\makecell{-}} & \multirow{2}{*}{\makecell{34}}  &  \multirow{2}{*}{\makecell{27}} & \multirow{2}{*}{\makecell{51}} & \multirow{2}{*}{\makecell{\xmark}} & \multirow{2}{*}{\makecell{\textbf{Sentence}}}
    \\
    \\
    \hline
    \multirow{3}{*}{\makecell{\textbf{LRW-ID} \\ \cite{kim2022speaker}}}
    & \textbf{Baseline} & Train & 17,560 &  155 & 500 & \multirow{3}{*}{\makecell{\cmark}}  & \multirow{3}{*}{\makecell{\textbf{Word}}}    
    \\ \cline{2-6}
    & \multirow{2}{*}{\makecell{\textbf{Adaptation}}} 
    & Train(Valid) & 20  &  9.6  & 500  \\
    &  & Test & 20  &  9.6  & 500 \\
    \hline
    \multirow{4}{*}{\makecell{\textbf{VoxLRS-SA}}}
    & \textbf{Baseline} & Train & 9,621  &  1,716 & 103,418 & \multirow{4}{*}{\makecell{\cmark}}  & \multirow{4}{*}{\makecell{\textbf{Sentence}}}   
    \\ \cline{2-6}
    & \multirow{3}{*}{\makecell{\textbf{Adaptation}}} 
    & Train & 20  &  18.4  & 10,912  \\
    & & Valid & 20  &  3.4  & 4,654  \\
    &  & Test & 20  &  2.0  & 3,460 \\
    \Xhline{3\arrayrulewidth}
  \end{tabular}}

  \caption{Comparison of our dataset with publicly available speaker adaptive lip reading datasets. This table outlines the differences in dataset configurations across various factors such as the number of words, pose, and background diversity.}
  \label{table:1}
\end{table*}


Although recent studies have advanced our understanding of speaker adaptation in lip reading, existing approaches suffer from several limitations. Firstly, previous works have focused solely on visual cues for speaker adaptation, overlooking the role of individual linguistic patterns. While language-level adaptation for target speakers has demonstrated effectiveness in the NLP domain \cite{kolar2007adaptation}, these effects remain unexplored in speaker-adaptive lip reading. Secondly, previous studies such as \cite{kim2022speaker} and \cite{he2024speaker} primarily utilize the GRID and LRW-ID datasets, which do not adequately capture the complexity of real-world scenarios. As shown in Table \ref{table:1}, these datasets especially lack pose diversity and have a limited vocabulary size.


To address these limitations, we propose a novel method to adapt the model to unseen target speakers in vision and language levels. This methodology comprises two primary components: 1) For vision level adaptation, the proposed method adjusts the pre-trained lip reading model to adapt to the lip appearance, movements, and speaking speed of the target unseen speaker while minimizing the number of training parameters. 2) For language level adaptation, we aim to adapt the pre-trained model to an individual's unique linguistic patterns such as frequent vocabulary choice at the language level, when predicting the target speaker's spoken language.

Specifically, we effectively design the vision-level adaptation by employing both padding prompts \cite{kim2022speaker} and LoRA \cite{he2024speaker} in the visual encoder, different from previous speaker adaptive lip reading methods solely relying on prompt or LoRA. Moreover, we focus on two components within the visual encoder. Initially, the padding prompt with LoRA is applied to the spatial encoding stage, which is crucial for adapting to the lip appearances of the target speaker. Subsequently, it is applied to the temporal encoding stage, enabling the model to better account for variations in lip movement speeds and styles specific to each speaker, thus enhancing the model's accuracy in capturing dynamic visual cues.  For language-level adaptation, we apply input prompt tuning to a decoder based on the fact that input-level modification of pre-trained models are effective for adapting them to different tasks \cite{brown2020language, zhou2022conditional}. With input prompt tuning, the decoder learns the probability of language modeling specific to the speaker when predicting the spoken language of the target speaker. 

To validate our proposed speaker adaptation method in real-world scenarios, we introduce a new dataset named VoxLRS-SA, derived from VoxCeleb2 \cite{chung2018voxceleb2} and LRS3 \cite{afouras2018lrs3} datasets. Together, these sources consist of about 1700 hours of YouTube videos, and as a result, VoxLRS-SA naturally captures diverse speaker poses and contains a rich vocabulary. Moreover, the VoxLRS-SA dataset combines the strengths of its parent datasets: LRS3, which includes text transcription labels but lacks specific speaker information, and VoxCeleb2, which provides detailed speaker information but does not include text transcriptions. To address these gaps, we utilize a pre-trained Automatic Speech Recognition (ASR) model \cite{radford2023whisper} to generate automatic text labels \cite{ma2023auto, yeo2024icassp} for the VoxCeleb2 dataset. Additionally, we employ a face recognition technology \cite{serengil2024lightface} to generate speaker information for the LRS3 dataset.

The key contributions of this paper are as follows: 1) To the best of our knowledge, this is the first speaker adaptive lip reading method leveraging multimodal information at vision and language levels and exploring its effectiveness in the wild sentence-level lip reading. Moreover, the proposed method effectively adapts to a target speaker by employing both prompts and LoRA approaches. 2) We introduce the VoxLRS-SA dataset to validate speaker adaptive lip reading methods in the wild, sentence-level lip reading in English.  It contains diverse poses and a vocabulary size of approximately 100K. 3) Through comprehensive experimental validation on the VoxLRS-SA dataset, we show that previous speaker-adaptive approaches also improve sentence-level performance in real-world settings. Furthermore, our newly proposed adaptation method surpasses previous methods in improving outcomes when tailored to specific target speakers.

\section{Related Work}
\subsection{Lip Reading}
Lip reading is the task of predicting spoken language by analyzing visual cues from lip movements. This technology has significantly advanced with the development of deep learning and the availability of large-scale audio-visual lip reading databases.

In word-level lip reading, the introduction of publicly available datasets such as LRW \cite{chung2017lip} and LRW-1000 \cite{yang2019lrw} has been crucial. Based on these databases, \cite{stafylakis2017combining} designed 3D convolutional layers and 2D ResNet architectures as the front-end, with LSTM networks as the back-end. Additionally, two-stream networks, which utilize both raw video and optical flow, have been proposed by \cite{xiao2020deformation, weng2019learning}. Some works \cite{kim2022distinguishing, yeo2023multi} have used audio signals to enhance the visual representations by learning visual-to-audio mapping in the memory network. 

In sentence-level lip reading, the release of the large-scale LRS2 \cite{afouras2018deep} and LRS3 \cite{afouras2018lrs3} databases has been important in English. With the development of these datasets, an end-to-end model trained using Connectionist Temporal Classification (CTC) loss \cite{graves2006connectionist} was proposed by \cite{assael2016lipnet}. Subsequently, \cite{son2017lip} developed a model utilizing the Seq2Seq architecture \cite{sutskever2014sequence}, and further architectural enhancements were achieved by \cite{afouras2018deep} with the implementation of the Transformer model \cite{vaswani2017attention}. Recently, state-of-the-art models \cite{ma2021end} have employed a joint CTC/Attention \cite{watanabe2017hybrid} loss with the Conformer \cite{gulati2020conformer} architecture. \cite{shi2022learning, haliassos2022jointly} utilized audio-visual data to train backbone models in a self-supervised manner and proved its effectiveness. In the context of other languages, LIP-RTVE dataset \cite{gimeno2023lip} provides a resource for automatic lip reading in continuous Spanish in the wild, featuring both speaker-independent and speaker-dependent partitions. Furthermore, with the development of self-supervised learning using audio-visual data, \cite{kim2024efficient, kim2023lip} have begun to explore lip reading technologies for multiple languages.

Although there has been great development in lip reading, these systems suffer from degraded performance when applied to unseen speakers not included in the training dataset, due to individual variations in lip appearances and movements. Therefore, it is necessary to explore speaker adaptation techniques to narrow the performance gap between seen and unseen speakers in lip reading. However, existing datasets \cite{kim2022speaker, cooke2006audio} for speaker-adaptive lip reading do not cover real-world scenarios for English, due to their limited vocabulary size and lack of pose diversity. To address this issue, we propose a novel dataset VoxLRS-SA, which contains about 100K vocabulary size and diverse pose videos. 


\begin{figure*}[t]
	\centering
	\centerline{\includegraphics[width=18cm]{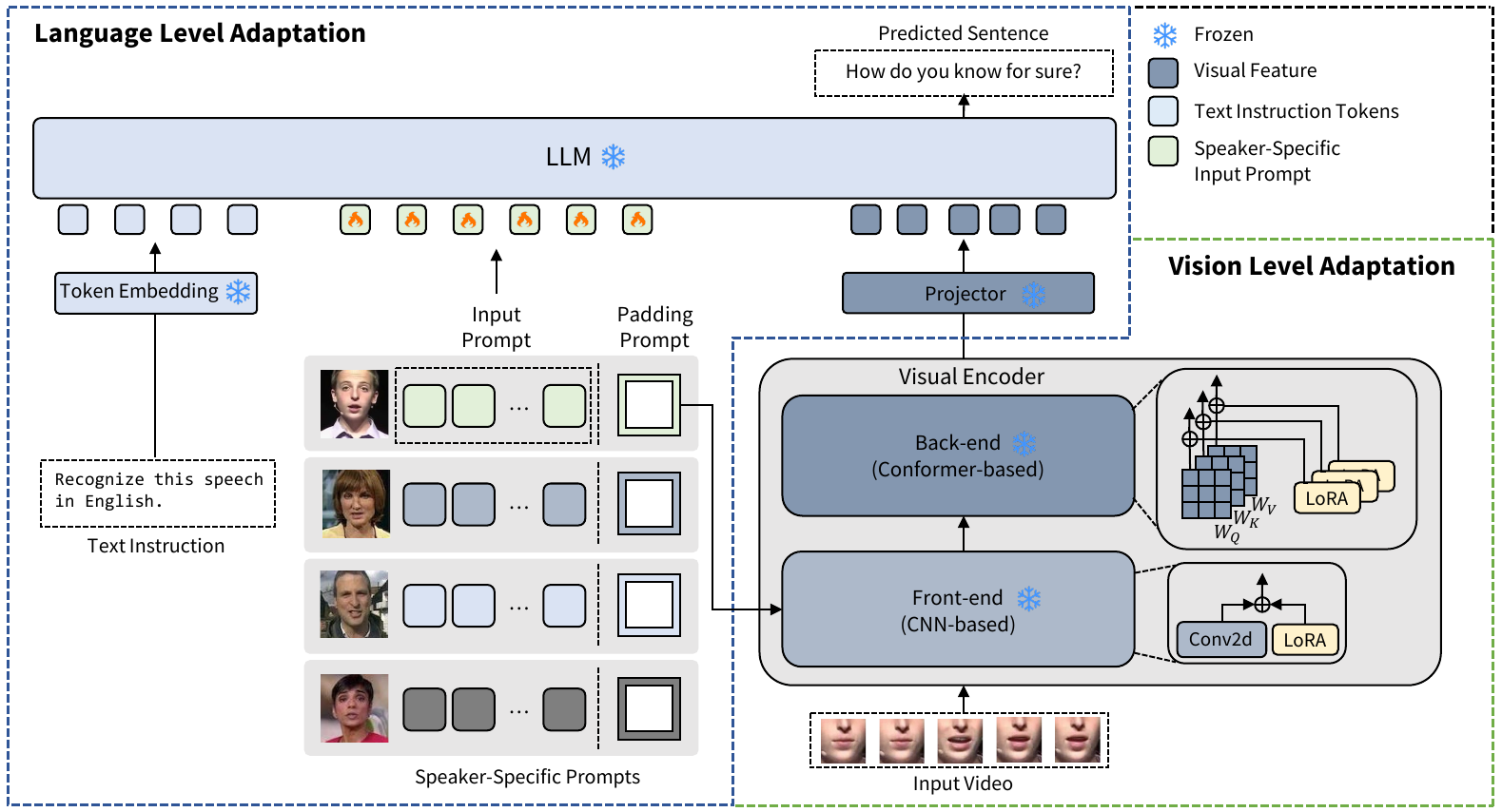}}

	\caption{Overview of the proposed lip reading model architecture illustrating dual adaptation strategies. Visual level adaptation employs padding prompts and LoRA at the front-end for unique lip features and at the back-end for lip movement speed and style. Language level adaptation uses input prompt tuning to learn the language modeling probability of each speaker.
	}
	\label{fig:1}
\end{figure*}


\subsection{Speaker Adaptation}
Speaker adaptation \cite{miao2015speaker} was primarily developed for Automatic Speech Recognition (ASR). An early work \cite{liao2013large} fine-tuned various parts of the network to adapt the model to a target speaker. Other studies \cite{li2010comparison, seide2011feature} suggested adding speaker-dependent layers to a pre-trained model. In \cite{swietojanski2014learning}, speaker-dependent vectors were added to all pre-trained hidden layers to adjust them for the test speaker. To provide additional inputs that vary by speaker, \cite{abdel2013rapid,abdel2013fast, xue2014fast} proposed speaker codes. Recently, works \cite{klejch2019speaker, huang2020using} have explored the utilization of meta-learning and speech synthesis for speaker adaptation, respectively.

In contrast to the efforts for speaker adaptation in ASR, only a few works handled speaker adaptation for lip reading. Motivated by the traditional speaker adaptation methods \cite{anastasakos1997speaker, gopinath1998maximum} of ASR, \cite{almajai2016improved} introduced utilize these works into lip reading. Recently,  \cite{kim2022speaker} proposed adapting the CNN layer of a visual encoder to the speaker's unique features, such as the shape of the lips, by using user-dependent padding. Moreover,  \cite{kim2023prompt,  he2024speaker} developed speaker adaptive lip reading by adapting the visual encoder to target the speaker at the input, spatial, and temporal levels. \cite{li2024generalizing, wu2024landmark} proposed using lip landmarks to mitigate the sensitivity of lip reading systems to individual variation information such as lip appearances.

Prior work has concentrated primarily on visual cues for speaker adaptation, overlooking the importance of individual linguistic patterns. \cite{kolar2007adaptation, guz2010segmentation} have utilized a speaker's lexical pattern for dialogue act and sentence segmentation tasks. However, leveraging linguistic cues has not yet been integrated into speaker-adaptive lip reading. By incorporating multimodal information, we improve speaker-specific adaptations, providing a more holistic approach to speaker adaptation.

\section{Method}
In this paper, the primary goal of our proposed method is to leverage individual characteristics at both the vision and language levels to improve the understanding of speaker-specific information, thereby offering a more comprehensive approach to speaker adaptation. 
 
Recent work \cite{yeo2024visual} proves that the LLM's context modeling capability is particularly effective in lip reading systems. Motivated by this, we employ this framework as our baseline lip reading model. As illustrated in Figure.\ref{fig:1}, the architecture of the lip reading model consists of two components, a visual encoder for modeling spatio-temporal information from lip movements, and an LLM as a decoder for predicting the sentence by taking visual features as input. To adapt the pre-trained lip reading model to target speakers, we propose a two-fold approach that addresses both the vision and language levels.

\subsection{Vision Level Adaptation}
The visual encoder of the pre-trained lip reading model consists of two parts: the front-end $\mathcal{F}$, which has a CNN-based architecture that encodes spatial information, and the back-end $\mathcal{B}$, which has a Conformer-based architecture that models temporal dependencies between lip sequences. Based on these two modules of the visual encoder, we specifically aim to adapt the visual encoder to lip appearances, movements, and the speaking speed of target speakers via vision-level adaptation.

Given the speaker-specific training dataset for speaker adaptation $\tau = \{(x_i, y_i)\}_{i=1}^N$, where $x_i \in \mathbb{R}^{T \times H \times W \times C}$ is the lip movements video of the target unseen speaker, $T$ is the frame length of the video, $y_i$ is the text transcriptions corresponding to its lip movements, and $N$ is the number of training samples. To adapt the appearances of the target speaker, a padding prompt \cite{kim2023prompt} and LoRA \cite{hu2021lora} are employed in the CNN layer of the front-end. These learnable parameters enable the front-end to adapt to the target speaker without any modifications to the pre-trained weights. Specifically, the padding prompt $p$ replaces the zero padding of the CNN layer. At the same time, the CNN layer is reconstructed with the low-rank decomposition matrices. With these two additional parameters, the spatial information encoding process can be formulated as follows: $f_s = \mathcal{F}_{\theta, p}(x_i)$, where $f_s \in \mathbb{R}^{T \times D}$ spatial feature of lip movements, $D$ is the embedding dimension of the front-end, ${\theta}$ indicates the weights of the front-end with the trainable parameters constructing the low-rank decomposition matrices, and $p$ indicates the replacement zero-padding with the padding prompts in the front-end.

For a better modeling of the speaker-specific temporal information such as speaking speed, the back-end needs to be adjusted. In order to effectively adjust without pre-trained weights of the back-end, LoRA is also utilized in the back-end. Specifically, LoRA is applied to the attention layer to learn the temporal dependencies between any two positions of speaker-specific lip movements. It can be formulated as follows:  $f_v = \mathcal{B}_{\theta}(f_s)$, where $f_v \in \mathbb{R}^{T \times D}$ is a visual feature containing spatio-temporal information of lip movements, and $\theta$ indicates the weights of the back-end with the trainable parameters constructing the low-rank decomposition matrices. The visual features encoded from the visual encoder are fed into the projector, to align it with the embedding space of the LLM. These aligned visual features are denoted as $f_v' \in \mathbb{R}^{T \times D_L}$, where $D_L$ is the embedding dimension of LLM. With the frozen LLM taking aligned visual features as input, the predicted sentence $\hat{y}_i$ is obtained. From this prediction, the cross entropy loss $\mathcal{L} = CE(\hat{y_i}, y_i)$ is calculated, and the visual encoder is optimized to adapt to the target speaker at the vision-level. 

\begin{table*}[t]
  \renewcommand{\arraystretch}{1.5}
  \renewcommand{\tabcolsep}{2mm}
  \centering
  \resizebox{0.999\linewidth}{!}{
  \begin{tabular}{cccccccccccc}
    \Xhline{3\arrayrulewidth}
     \multirow{2}{*}{\textbf{Split}} &  \multirow{2}{*}{\makecell{\textbf{Data} \\ \textbf{Information}}} & \multicolumn{10}{c}{\textbf{Speaker Number}}
    \\
    \cline{3-12}
    & & \textbf{S1 $|$ S2} & \textbf{S3 $|$ S4} & \textbf{S5 $|$ S6} & \textbf{S7 $|$ S8} & \textbf{S9 $|$ S10} & \textbf{S11 $|$ S12} & \textbf{S13 $|$ S14} & \textbf{S15 $|$ S16} & \textbf{S17 $|$ S18} & \textbf{S19 $|$ S20}
    \\ \hline
    \multirow{3}{*}{\textbf{Train}} 
    & \# videos & 280 $|$ 261 & 281 $|$ 291 &  268 $|$ 280 & 297 $|$ 283 & 294 $|$ 253 & 308 $|$ 316 & 286 $|$ 304 & 279 $|$ 315 & 302 $|$ 322 & 305 $|$ 317
    \\ & Duration(min) & 45.05 $|$ 45.06 & 45.09 $|$ 45.13 & 45.08$|$ 45.12 & 45.10 $|$ 45.05 & 45.12 $|$ 45.26 & 45.05 $|$ 45.00 & 45.07 $|$ 45.01 & 45.10 $|$ 44.99 & 45.00 $|$ 45.05 & 45.09 $|$ 45.06 
    \\ & \# words & 1585 $|$ 1489 & 1796 $|$ 1645 & 1433 $|$ 1409 & 1535 $|$ 1671 & 1466 $|$ 1614 & 1535 $|$ 1400 & 1637 $|$ 1394 & 1344 $|$ 1560 & 1548 $|$ 1409 & 1485 $|$ 1423
    \\
    \hline
    \multirow{4}{*}{\textbf{Valid}} 
    & \# videos & 94 $|$ 85 & 78 $|$ 83 &  74 $|$ 74 & 68 $|$ 66 & 74 $|$ 63 & 71 $|$ 61 & 68 $|$ 62 & 52 $|$ 60 & 47 $|$ 42 & 38 $|$ 44
    \\ & Duration(min) & 14.07 $|$ 13.87 & 13.01 $|$ 12.01 & 11.83 $|$ 11.46 & 11.51 $|$ 11.30 & 11.16 $|$ 10.77 & 10.31 $|$ 9.66 & 9.46 $|$ 9.22 & 8.5 $|$ 8.27 & 7.66 $|$ 6.00 & 5.92 $|$ 5.85 
    \\ & \# words & 774 $|$ 720 & 800 $|$ 731 & 657 $|$ 641 & 674 $|$ 726 & 655 $|$ 692 & 591 $|$ 569 & 619 $|$ 548 & 483 $|$ 540 & 507 $|$ 444 & 441 $|$ 392
    \\ & \makecell{overlap ratio} & 0.33 $|$ 0.27 & 0.26 $|$ 0.25 & 0.27 $|$ 0.27 & 0.28 $|$ 0.25 & 0.28 $|$ 0.27 & 0.24 $|$ 0.28 & 0.24 $|$ 0.25 & 0.23 $|$ 0.21 & 0.22 $|$ 0.21 & 0.19 $|$ 0.20
    \\
    \hline
    \multirow{4}{*}{\textbf{Test}} 
    & \# videos & 41 $|$ 35 & 35 $|$ 41 &  35 $|$ 43 & 33 $|$ 50 & 34 $|$ 41 & 37 $|$ 41 & 40 $|$ 34 & 44 $|$ 43 & 35 $|$ 42 & 39 $|$ 46
    \\ & Duration(min) & 6.12 $|$ 6.10 & 6.15 $|$ 6.10 & 6.04 $|$ 6.09 & 6.08 $|$ 6.12 & 6.09 $|$ 6.08 & 6.12 $|$ 6.10 & 6.16 $|$ 6.10 & 6.16 $|$ 6.15 & 6.15 $|$ 6.15 & 6.11 $|$ 6.15 
    \\ & \# words & 461 $|$ 425 & 507 $|$ 459 & 427 $|$ 438 & 425 $|$ 419 & 419 $|$ 491 & 396 $|$ 468 & 499 $|$ 396 & 368 $|$ 425 & 402 $|$ 438 & 473 $|$ 388
    \\ & \makecell{overlap ratio} & 0.18 $|$ 0.18 & 0.18 $|$ 0.17 & 0.21 $|$ 0.20 & 0.17 $|$ 0.18 & 0.20 $|$ 0.19 & 0.18 $|$ 0.20 & 0.16 $|$ 0.19 & 0.19 $|$ 0.18 & 0.19 $|$ 0.21 & 0.20 $|$ 0.19
    \\
    \hline

  \end{tabular}}

  \caption{Detailed data information of 20 test speakers in VoxLRS-SA}
  \label{table:2}
\end{table*}

\subsection{Language Level Adaptation}
Speaker-specific characteristics encompass not only visual cues such as the appearance of lips but also linguistic patterns, including preferred vocabulary choices. Since individuals exhibit unique linguistic patterns, the likelihood of specific sequences of words occurring varies from person to person. To learn these speaker-specific linguistic patterns, our approach utilizes an input prompt tuning motivated by \cite{brown2020language, zhou2022conditional, lester2021power}. 

Specifically, given that the visual feature $f_v'$ is already aligned with the embedding space of the LLM, the input prompts, as trainable parameters, are concatenated with these features along the temporal dimension and then fed into the LLM as input. Consequently, the input prompt is designed to have the same dimension as that of the LLM embedding space dimension and can be denoted as $p_i \in \mathbb{R}^{N_p \times D_L}$, where $N_p$ is the length of the prompts. With the self-attention layers in the LLM and model temporal dependencies between input prompts and visual features, these prompts can influence speaker-specific sentence predictions. It can be formulated as follows: $\hat{y}_i =\mathcal{D}(p_i \oplus f_v')$, where the $\oplus$ is the concatenate function and $\mathcal{D}$  represents the LLM functioning as the decoder for sentence prediction. In addition, the input prompts are trained to predict more accurate language modeling probabilities for specific speakers using target speaker data. This objective is achieved by minimizing the cross-entropy loss, which can be formulated as $\mathcal{L} = CE(\hat{y_i}, y_i)$, which optimizes the input prompt for language-level adaptation.

\subsection{VoxLRS-SA}
To validate the effectiveness of the proposed method in real-world scenarios for English, we develop the VoxLRS-SA dataset based on LRS3 and VoxCeleb2 datasets. While the widely used LRS3 dataset for sentence-level lip reading in the wild includes human-annotated transcription labels, it does not contain Speaker ID information for each video. In contrast, the VoxCeleb2 dataset includes Speaker ID information but does not provide text transcriptions. To address this gap, we generate the speaker ID information for the LRS3 dataset by using face verification \cite{kim2022speaker} and make pseudo text labels for the VoxCeleb2 dataset by employing a pre-trained ASR model \cite{ma2023auto, yeo2024icassp}. 

\subsubsection{Automatic Labels in VoxCeleb2}
Recent works \cite{ma2023auto, yeo2024icassp} have demonstrated that a lip reading model trained using automatically generated transcriptions through a pre-trained ASR model performs similarly to one trained with human-annotated labels. Motivated by these works, we utilize the Whisper \cite{radford2023whisper} ASR model to generate automatic labels for the videos classified as the English portion of the VoxCeleb2 dataset. Then, we choose 20 speakers who have more than 50 minutes of video content to construct the test and adaptation (or validation) sets. Additionally, we manually clean the text transcriptions of the test set to accurately evaluate the effectiveness of the speaker adaptation method.

\subsubsection{Speaker ID information annotation}
Since the pseudo-labels are generated using the pre-trained ASR model, their transcription quality may be worse than human-annotated labels. To mitigate this issue, we combine VoxCeleb2 with the LRS3 dataset, which contains manually annotated transcriptions corresponding to 433 hours of videos.  

Since the LRS3 dataset does not include speaker ID information, we first generate the speaker ID information for both datasets using a method similar to the one described in \cite{kim2022speaker}, employing a face recognition model. Since there may be overlapping speakers between the two datasets, we take the following approach: For each video in LRS3 and VoxCeleb2, we divide the video into three equal segments and extract one face image from each segment. Then, we employ the Facenet \cite{serengil2024lightface} face recognition model to encode the facial images into features of speaker information. To identify overlapping speakers between the two datasets, we compare the feature vectors from LRS3 and VoxCeleb2. This comparison is done using a similarity measure, such as cosine similarity, to identify the overlapping speakers. 

\section{Experimental Setup}
\subsection{Dataset}
\noindent \textbf{VoxLRS-SA}
 is divided into two parts corresponding to each purpose. Firstly, the Baseline part of VoxLRS-SA is designed to train the baseline lip reading model. This training dataset is composed of  1.7K hours of video data from about 9K speakers. Secondly, the Adaptation part is proposed to adapt the pre-trained lip reading model to the target speaker and evaluate the effectiveness of the adaptation method. For this purpose, we construct the Adaptation part with 20 speakers, which do not overlap with the 9K speakers of the Baseline part. The \textit{train} split contains video data varying from 45 minutes per speaker, and the \textit{test} and \textit{val} sets contain about 5 minutes of data for each of the 20 speakers. The detailed data information is shown in Table \ref{table:2}.

\noindent \textbf{VoxCeleb2}
\cite{chung2018voxceleb2} is a speaker verification dataset. It contains only speaker information for over 6,000 speakers and 2,442 hours of multilingual videos. By referring to the \cite{shi2022learning}, we utilize an English portion, which amounts to 1,326 hours. 

\noindent \textbf{LRS3}
\cite{afouras2018lrs3} is a widely used sentence-level lip reading dataset comprising 433 hours of English talking face videos and human-annotated transcriptions. These talking face videos are collected from TED and TEDx talks, thus contain diverse poses.

\subsection{Implementation details}
\textbf{Preprocessing} By following a preprocessing process \cite{ma2023auto}, we crop the mouth region from talking face videos into a size of 96 pixels in both width and height. We employ RetinaFace \cite{deng2020retinaface} to detect facial landmarks and crop the mouth region based on these landmarks. For data augmentation \cite{shi2022learning}, horizontally flipping and randomly cropping are used.

\noindent \textbf{Architecture}
The design of our visual encoder incorporates a conformer architecture, configured with an embedding dimension of 768, 12 attention heads, a feed-forward dimension of 3072, and 12 blocks. The configuration of the transformer decoder, which is used for pre-training the visual encoder, includes an embedding dimension of 768, 8 attention heads, a feed-forward dimension of 3072, and 9 blocks. We employ the LLaMA3-8B \cite{touvron2023llama} model as the LLM for our proposed method. 

\noindent \textbf{Baseline Lip Reading Model}
We initialize the visual encoder with weights from a pre-trained lip reading model trained on the LRS3 dataset \cite{ma2023auto}. The visual encoder is fine-tuned alongside a transformer decoder from scratch on the Baseline part of the VoxLRS-SA dataset. During the first stage of training, we employ a tri-stage learning rate with a peak learning rate of 1e-3, warmup steps of 10K, decay steps of 20K, and utilize 8 NVIDIA RTX 3090 GPUs with a maximum frame count of 1800 and gradient accumulation set to 4. In a further step of our experiment, we replace the transformer decoder with the frozen LLaMA3-8B model, continuing the training on the same datasets. In this second stage of training, we adopt a cosine learning rate strategy with a learning rate of 5e-5, total training steps of 30K, and a warmup period of 0.5K steps, using a batch size of 1 and increasing the gradient accumulation to 8.

\noindent \textbf{Speaker-Adaptive Lip Reading Model}
For vision-level adaptation, a number of training steps is set to 300, and use a cosine learning rate scheduler without warmup. The learning rate begins at 1e-4 and gradually decreases to a minimum of 1e-5 over 5000 update periods. The LoRA configuration includes a rank of 8 and a scaling factor of 16. The target modules for LoRA adaptation are the convolutional layer, a query, a key, and a value in the self-attention layer. For language-level adaptation, the visual encoder is kept frozen to focus adaptation efforts on the language model. The LoRA configuration is applied to the attention layers of the LLM, specifically targeting the query, key, and value components. The number of training updates is set to 70.  

\begin{table}[t]
\renewcommand{\arraystretch}{1.5}
\renewcommand{\tabcolsep}{5.0mm}
  \centering
  \resizebox{0.999\linewidth}{!}{
  \begin{tabular}{cccc}
    \Xhline{3\arrayrulewidth}
    \multirow{2}{*}{\textbf{Method}} & \multicolumn{2}{c}{\textbf{Trainable Params}} & \multirow{2}{*}{\textbf{WER(\%)}} 
    \\
    \cline{2-3}
    & \makecell{\textbf{Encoder}} & \makecell{\textbf{Decoder}} & 
    \\
    \hline 
     Baseline & 181.3M$^*$ & 8B$^*$ & 47.3 \\
     \hdashline
     \multicolumn{4}{c}{\textbf{Vision-Level Adaptation}} \\
     V LoRA & +0.6M & - & 42.5  \\
     Padding Prompt & +0.1M & - & 42.9  \\
     Fintune-F & 11.2M & - & 42.8  \\
     Fintune-B & 170.1M & - & 40.2  \\
    Finetune-F\&B & 181.3M & - & 40.0 \\
     Ours & +0.7M & - & 41.5 \\
    \hdashline
    \multicolumn{4}{c}{\textbf{Language-Level Adaptation}} \\
     L LoRA & - & +4.7M & 44.7 \\
     Ours & - & +0.04M  & 44.1 \\
    \hdashline
    \multicolumn{4}{c}{\textbf{Vision-and-Language-Level Adaptation}} \\
    Finetune-F\&B & 181.3M & +0.04M & 39.6 \\
    Ours & +0.7M & +0.04M & 40.9 \\
    \Xhline{3\arrayrulewidth}
  \end{tabular}}

  \caption{This table compares the WER performance ($\downarrow$) for various speaker-adaptive training strategies, detailing changes in trainable parameters for the encoder and decoder. $^*$ indicates the number of total parameters.}

  \label{table:3}

\end{table}

\section{Experimental Results}

\subsection{Baselines for comparisons}
Since this is the first work to explore the effectiveness of speaker-adaptive lip reading at the wild sentence level in English, we have established comparison methods including previous word-level speaker-adaptive methods and full fine-tuning methods to validate the effectiveness of the proposed method.

\noindent \textbf{LoRA} \cite{hu2021lora} is a currently widely used technique for computationally efficient fine-tuning. We set three types of adaptation methods based on LoRA. The V LoRA and L LoRA are adapting the visual encoder at vision level adaptation and LLM at language-level adaptation, respectively.

\noindent \textbf{Padding Prompt} \cite{kim2022speaker} utilize additional speaker-specific input in the convolution layer of the visual encoder. It enables the speaker-specific input to memorize the visual characteristics of the target speaker.

\noindent \textbf{Finetune} By following \cite{kim2023prompt}, we finetune three parts. Finetune-F and -B are fine-tuning the front- and back-ends of the visual encoder, respectively. Finetune-F\&B is fine-tuning the whole part of the visual encoder.

\subsection{Comparison with the previous method at Vision and Language Levels Adaptation}

Table \ref{table:3}. provides details of experiments conducted to evaluate different training strategies in speaker adaptive lip reading. The baseline method consists of a considerable number of parameters, with 181.3 million in the encoder and 8 billion in the decoder, resulting in a Word Error Rate (WER) of 47.3\%. 

\begin{table*}[t]
  \renewcommand{\arraystretch}{1.5}
  \renewcommand{\tabcolsep}{2mm}
  \centering
  \resizebox{0.999\linewidth}{!}{
  \begin{tabular}{cccccccccccc}
    \Xhline{3\arrayrulewidth}
     \multirow{2}{*}{\makecell{\textbf{Adapt} \\ \textbf{Min}}} & \multicolumn{10}{c}{\textbf{Speaker Number}} & \multirow{2}{*}{\textbf{Mean}($\downarrow$)}
    \\
    \cline{2-11}
    & \textbf{S1 $|$ S2} & \textbf{S3 $|$ S4} & \textbf{S5 $|$ S6} & \textbf{S7 $|$ S8} & \textbf{S9 $|$ S10} & \textbf{S11 $|$ S12} & \textbf{S13 $|$ S14} & \textbf{S15 $|$ S16} & \textbf{S17 $|$ S18} & \textbf{S19 $|$ S20}
    \\ \hline 
    \textbf{Baseline} & 42.2 $|$ 43.1 & 59.5 $|$ 48.2 &  32.7 $|$ 83.8 & 41.9 $|$ 80.4 & 32.3 $|$ 64.9 & 49.3 $|$ 37.0 & 66.9 $|$ 31.8 & 33.5 $|$ 48.2 & 53.3 $|$ 31.0 & 29.6 $|$ 35.9 & 47.3
    \\
    \textbf{1 min} & 43.7 $|$ 44.9 & 53.0 $|$ 44.1 &  39.3 $|$ 71.5 & 44.1 $|$ 76.4 & 31.5 $|$ 58.4 & 47.5 $|$ 36.6 & 57.3 $|$ 29.4 & 37.8 $|$ 48.0 & 56.0 $|$ 31.7 & 36.6 $|$ 35.7 & 46.2
    \\
    \textbf{5 min} & 40.9 $|$ 42.7 & 50.5 $|$ 41.3 &  30.9 $|$ 68.5 & 43.9 $|$ 70.5 & 29.8 $|$ 56.5 & 44.1 $|$ 37.7 & 57.5 $|$ 28.4 & 33.4 $|$ 45.2 & 52.6 $|$ 28.8 & 29.4 $|$ 34.4 & 43.4
    \\
    \textbf{15 min} & 39.6 $|$ 39.4 & 50.8 $|$ 42.0 &  28.3 $|$ 69.9 & 41.8 $|$ 71.1 & 29.1 $|$ 55.2 & 45.7 $|$ 36.7 & 53.9 $|$ 27.5 & 32.0 $|$ 44.4 & 51.4 $|$ 28.3 & 29.0 $|$ 33.1 & 42.5
    \\
    \textbf{30 min} & 37.9 $|$ 38.7 & 51.9 $|$ 41.7 & 26.8 $|$ 67.9 & 40.2 $|$ 66.7 & 29.7 $|$ 55.9 & 42.1 $|$ 34.8 & 55.1 $|$ 28.3 & 32.8 $|$ 43.9 & 48.1 $|$ 26.0 & 28.2 $|$ 32.2 & 41.4
    \\
    \textbf{45 min} & 35.4 $|$ 38.1 & 48.6 $|$ 39.6 & 27.5 $|$ 64.7 & 41.2 $|$ 67.1 & 29.9 $|$ 55.0 & 44.1 $|$ 34.0 & 54.1 $|$ 27.5 & 31.6 $|$ 43.2 & 50.2 $|$ 27.1 & 28.5 $|$ 30.7 & 40.9
    \\
    \Xhline{3\arrayrulewidth}

  \end{tabular}}

  \caption{WER performance ($\downarrow$) across different adaptation durations: This table shows the WER for 20 speakers, across different adaptation durations from 1 up to 45 minutes. Each duration denotes the total length of the video sample used for adaptation.}
  \label{table:4}

\end{table*}

In the vision-level adaptation, we compare the proposed method with V LoRA, Padding Prompt, and three types of full fine-tuning to validate the effectiveness. V LoRA and Padding Prompt added a small amount of trainable parameters (0.6M and 0.1M, respectively) but achieved improvements in WER, reducing it to 42.5\% and 42.9\%, respectively. The full-fine-tuning methods Finetune F, B, and F\&B, obtained substantial reductions, with results of 42.8\%, 40.2\%, and 40.0\%, respectively. While these methods adapting the back-end modules show superior performances, they require training over 170M parameters. In contrast, our proposed method achieves a WER of 41.5\% by training only 0.7M parameters. It is worth noting that the proposed method outperforms the Finetune-F method, which utilizes 11.2M parameters.

In the language-level adaptation, we have kept the parameters of the visual encoder frozen to focus on verifying the effectiveness of adapting speaker-specific language information. We compare the proposed method, which utilizes input prompt tuning, with the LoRA-based approach. By employing the LoRA approach in the LLM for language level adaptation, we obtain a WER of 44.7\%. With the proposed method, a WER of 44.1\% is achieved by using only 0.04M learnable parameters. Despite the proposed method utilizing fewer parameters compared to the LoRA approach, it shows better performance.

In the vision-and-language level adaptation, we compare the proposed method with the Finetune F\&B, which shows the best performance in the vision-level adaptation. For a fair comparison, we also apply the language-level adaptation on Finetune F\&B using our input prompt. The results are shown in the lower part of Table \ref{table:3}. With the Finetune F\&B adaptation method, a WER of 39.6\% is achieved, which is the best performance in these experiments. The proposed method achieves a marginally higher WER of 40.9\% than Finetune F\&B by utilizing adaptations on both vision and language levels for the target speaker. Despite showing marginally lower performance compared to Finetune F\&B, the proposed method has the advantage of using only 0.74M parameters, which is less than 0.1\% of the number of parameters of Finetune F\&B.

\subsection{Analysis of Adaptation Duration Effects on Lip Reading Performances}
One of the challenges in speaker-adaptive lip reading is that it is not easy to collect sufficient data for the target speaker. Therefore, we validate the effectiveness of the proposed method by using a small amount of adaptation data corresponding to 1 and 5 minutes. Moreover, in order to verify the effectiveness according to the amount of adaptation data, we additionally conduct experiments by utilizing 15, 30, and 45 minutes of adaptation data.

Table \ref{table:4} illustrates the effect on target speaker lip reading performance according to variations of the amount of speaker-specific data, across 20 speakers. The WER performances of a baseline across 20 speakers range widely, with the lowest WER observed at 29.6 and the highest at 83.8, demonstrating variability in pre-trained baseline lip reading model accuracy across different speakers. The overall mean WER for the baseline is calculated at 47.3.

When we apply the proposed method across 20 speakers, the average performance is increased in all settings, regardless of the amount of adaptation data. Specifically, the speaker-adaptive lip reading model trained using 1-minute speaker-specific data achieves an average of 46.2\% WER, which is 1.2\% better than the baseline.  However, the performance of some speakers is worse than the baseline. This result is different from \cite{kim2022speaker}, which shows that 1 minute of adaptation data is enough to improve the performance of all speakers in word-level lip reading. In sentence-level speaker adaptive lip reading, it seems that there is a need to be over 1 minute of speaker-specific data, to adapt the pre-trained model to the target speaker.

By analyzing the results when using 5 minutes of speaker-specific data, the performance of the adaptive lip reading model outperforms that of the baseline for 19 out of 20 speakers. In other words, utilizing 5 minutes of speaker-specific data is enough to adapt the pre-trained model to target speakers. The adapted model achieves an average WER of 43.4\%, which is a 3.9\% improvement to the baseline model. Overall, employing more speaker-specific training data up to 45 minutes improves the average WER without saturation of performances, as seen in the average WERs of 42.5\%, 41.4\%, and 40.9\% for 15, 30, and 45 minutes, respectively.

\section{Conclusion}
This study has introduced a novel speaker-adaptive lip reading method that can be adapted to specific target speakers at both the vision and language levels. Specifically, we have adapted the visual encoder to the visual information of the target speaker, such as lip appearances and speaking speed, by using padding prompts and LoRA. Moreover, we have also adapted the LLM to speaker-specific language information by utilizing input prompt tuning for language-level adaptation. To validate the effectiveness of the proposed method, we have introduced a novel VoxLRS-SA dataset that expanded the vocabulary size and poses variations, compared to previous speaker-adaptive datasets in English. The experimental results have shown the necessity of speaker-adaptive lip reading in real-world scenarios. Moreover, our method not only improves sentence-level lip reading by using only a small amount of data but also outperforms existing speaker-adaptive methods. These findings suggest that incorporating both visual adaptation and language-specific characteristics of speakers can substantially benefit lip reading technologies.

\section{Acknowledgments}
This work was partly supported by two funds: the National Research Foundation of Korea (NRF) grant funded by the Korea government (MSIT) (No. NRF-2022R1A2C2005529)
 and  IITP grant funded by the Korea government(MSIT) (No.2020-0-00004, Development of Previsional Intelligence based on Long-Term Visual Memory Network)

\bigskip

\bibliography{aaai25}

\appendix

\section{Additional Results}

\subsection{Selecting Optimal Weights for Lora}
We have conducted eight additional experiments to assess the effectiveness of LoRA across different weight types in vision-level adaptation, as detailed in Table \ref{table:5}. The components $W_c$, $W_q$, $W_k$, and $W_v$ represent the LoRA weights for convolution, query, key, and value, respectively, with performance evaluated using WER. Among them, $W_v$ showed the most significant impact when applied LoRA to a single weight type, reducing the WER to 39.6\%. Applying LoRA to combined weights, the configuration of $W_q$, $W_k$, $W_v$ achieved a WER of 38.5\%. Extending this adaptation to include $W_c$, resulting in the configuration $W_c$, $W_q$, $W_k$, $W_v$, achieved a WER of 38.6\%, similar to the previous configuration. Furthermore, configurations at different ranks—8, 4, and 2—have achieved WERs of 38.6\%, 38.9\%, and 38.4\%, respectively. These results are significantly better compared to the baseline WER of 42.2\%.

\subsection{Failure Cases}
We have evaluated failure cases, where our proposed method shows higher WER than the baseline, and noticed two general patterns: misinterpretation of context and substitution errors. Our proposed method may distort the intended context by focusing on individual speaking styles rather than the actual spoken words. For example, when the ground truth is "but donald trump had nothing but praise for theresa may in this interview wants to see", our proposed method predicts as "that technology has nothing to do with the rise of making this even worse". Also, even with the speaker-adaptive approach, there are instances where our method substitutes words with similar-sounding words. For example, the ground truth is "and likes prime minister is people know that he is not interested in the office for what it can give him he's interested because of public service and he thinks two terms at the end of those two terms", and the prediction is “that line's premise is people know that he's not interested in the office for what he'll give him he's interested in the plans of public service and he says two times and the end of those two times”. The substitution of "likes" with "that line’s," "prime minister" with "premise," and "he'll" with "he'll give him" demonstrates that the method occasionally struggles to capture intended words.

\section{Limitation}
While we have utilized efficient adaptation methods such as LoRA and prompt tuning, which require only minimal adjustments to the pre-trained model's parameters, these methods still necessitate further training. Moreover, although our proposed method showed promising results when adapting the model using only 5 minutes of speaker-specific data, obtaining such data from a broad range of speakers is not easy. Therefore, as a future direction, we expect to explore methods that can effectively learn from even less data to mitigate this limitation, thus improving the adaptability and practicality of the model.

\begin{table}[t]
\renewcommand{\arraystretch}{1.5}
\renewcommand{\tabcolsep}{5.0mm}
  \centering
  \resizebox{0.999\linewidth}{!}{
  \begin{tabular}{ccc}
    \Xhline{3\arrayrulewidth}
    \textbf{Weight Type} & \textbf{Rank} $r$ & \textbf{WER(\%)} 
    \\
    \hline 
     Baseline & - & 42.2 \\
     \hdashline
     $W_c$ & 8 & 40.8   \\
     $W_q$ & 8 & 41.0   \\
     $W_k$ & 8 & 40.6  \\
     $W_v$ & 8 & 39.6   \\
     $W_q$, $W_k$, $W_v$ & 8 & 38.5  \\
     $W_c$, $W_q$, $W_k$, $W_v$ & 8 & 38.6  \\
     $W_c$, $W_q$, $W_k$, $W_v$ & 4 & 38.9  \\
     $W_c$, $W_q$, $W_k$, $W_v$ & 2 & 38.4  \\
    \Xhline{3\arrayrulewidth}
  \end{tabular}}

  \caption{Comparative analysis of WER performance ($\downarrow$) across various LoRA weight configurations and ranks in vision-level adaptation, tested on the S1 speaker from the VoxLRS-SA dataset.}

  \label{table:5}

\end{table}


\begin{figure}[t]
	\centering
    \centerline{\includegraphics[width=9cm]{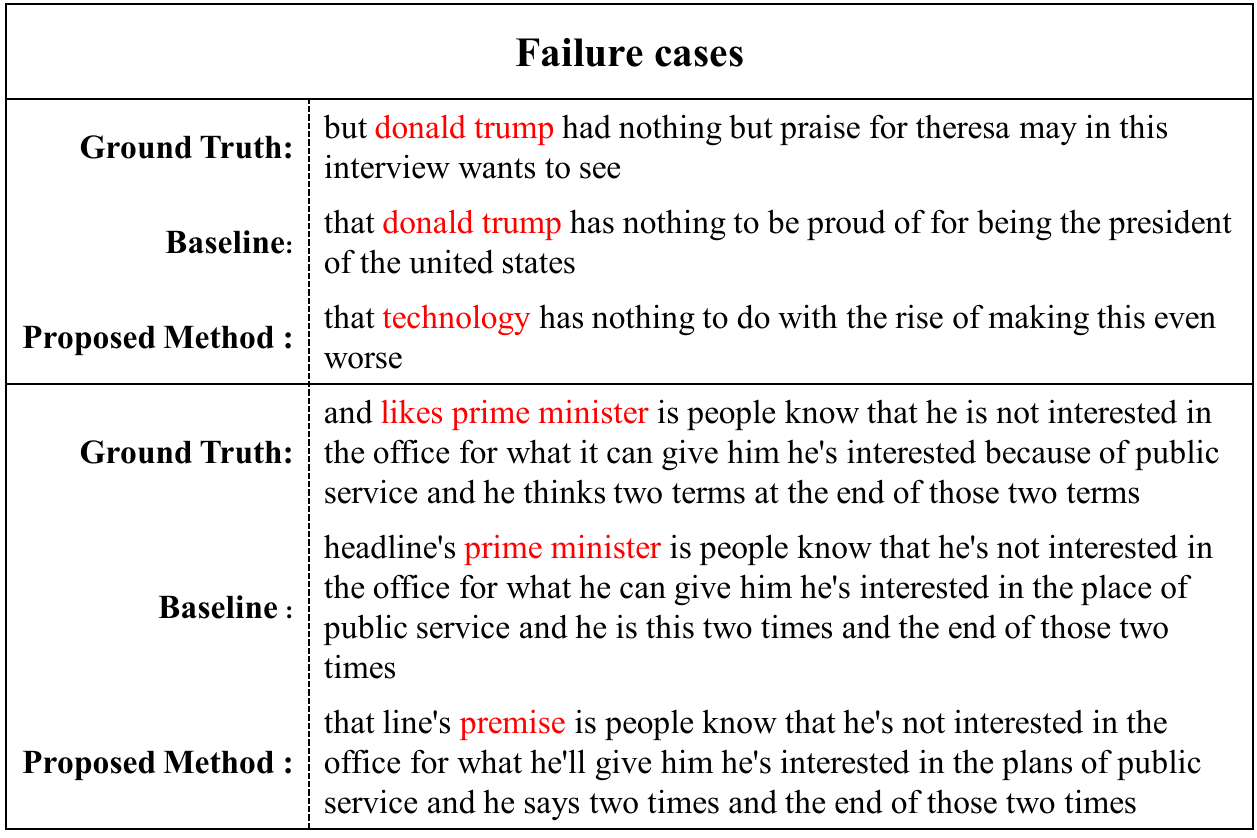}}
	\caption{Examples of failure cases highlighting misinterpretation of context and substitution errors in predictions}
	\label{sup_fig:1}
\end{figure}


\end{document}